\pdfoutput=1

\documentclass[11pt]{article}

\usepackage[]{EMNLP2022}

\usepackage{times}
\usepackage{latexsym}
\usepackage{microtype}
\usepackage{amsmath}
\usepackage{amssymb}
\usepackage{booktabs}
\usepackage{multicol}
\usepackage{multirow}
\usepackage{graphicx}  
\usepackage{float}  
\usepackage{subfigure}  
\usepackage{algorithm}
\usepackage{algorithmic}
\usepackage{makecell}
\DeclareMathOperator*{\argmax}{arg\,max}

\newcolumntype{L}[1]{>{\raggedright\let\newline\\\arraybackslash\hspace{0pt}}m{#1}}
\newcolumntype{C}[1]{>{\centering\let\newline\\\arraybackslash\hspace{0pt}}m{#1}}
\newcolumntype{R}[1]{>{\raggedleft\let\newline\\\arraybackslash\hspace{0pt}}m{#1}}
\newcommand{\secref}[1]{\S\ref{#1}}
\usepackage[T1]{fontenc}

\usepackage[utf8]{inputenc}

\usepackage{microtype}

\usepackage{inconsolata}

\title{Estimating Soft Labels for Out-of-Domain Intent Detection}

\author{
Hao Lang \thanks{\quad Equal contribution.} \quad 
Yinhe Zheng \footnotemark[1] \ \thanks{\quad Corresponding author.} \quad 
Jian Sun \quad 
Fei Huang \quad 
Luo Si \quad 
Yongbin Li \footnotemark[2] \\
  Alibaba Group \\
    \texttt{\{hao.lang, jian.sun, f.huang, luo.si,shuide.lyb\}@alibaba-inc.com}, \ \\  \texttt{zhengyinhe1@163.com}  \\}

\begin{document}
\maketitle
\begin{abstract}

Out-of-Domain (OOD) intent detection is important for practical dialog systems. To alleviate the issue of lacking OOD training samples, some works propose synthesizing pseudo OOD samples and directly assigning one-hot OOD labels to these pseudo samples. However, these one-hot labels introduce noises to the training process because some ``hard'' pseudo OOD samples may coincide with In-Domain (IND) intents. In this paper, we propose an \underline{a}daptive \underline{so}ft pse\underline{u}do \underline{l}abeling (ASoul) method that can estimate soft labels for pseudo OOD samples when training OOD detectors. Semantic connections between pseudo OOD samples and IND intents are captured using an embedding graph. A co-training framework is further introduced to produce resulting soft labels following the \emph{smoothness assumption}, i.e., close samples are likely to have similar labels. Extensive experiments on three benchmark datasets show that ASoul consistently improves the OOD detection performance and outperforms various competitive baselines.
\end{abstract}

\section{Introduction}

Intent detection is essential for dialogue systems, and current methods usually achieve high performance under the \textit{closed-world assumption} \cite{shu2017doc}, i.e., data distributions are static, and only a fixed set of intents are considered. However, such an assumption may not be valid in practice, where we usually confront an \textit{open-world} \citep{fei2016breaking}, i.e., unknown intents that are not trained may emerge. It is necessary to equip dialogue systems with Out-of-Domain (OOD) detection abilities so that they can accurately classify known In-Domain (IND) intents while rejecting unknown OOD intents \cite{yan-etal-2020-unknown,shen2021enhancing}.

A major challenge for OOD detection is the lack of OOD samples \cite{xu2020deep}. In most applications, it is hard, if not impossible, to collect OOD samples from the test distribution before training \cite{du2021vos}. To tackle this issue, various studies try to synthesize pseudo OOD samples in the training process. Existing methods include distorting IND samples \cite{choi2021outflip,shu2021odist,ouyang2021energy}, using generative models \cite{ryu-etal-2018-domain,zheng2020out}, or even mixing-up IND features \cite{zhou2021learning,zhan2021out}. Promising results are reported by training a ($k+1$)-way classifier ($k$ IND classes + 1 OOD class) using these pseudo OOD samples \cite{geng2020recent}. This classifier can classify IND intents while detecting OOD intent since inputs that fall into the OOD class are regarded as OOD inputs.

\begin{figure}[t]
  \centering
  \includegraphics[width=170px]{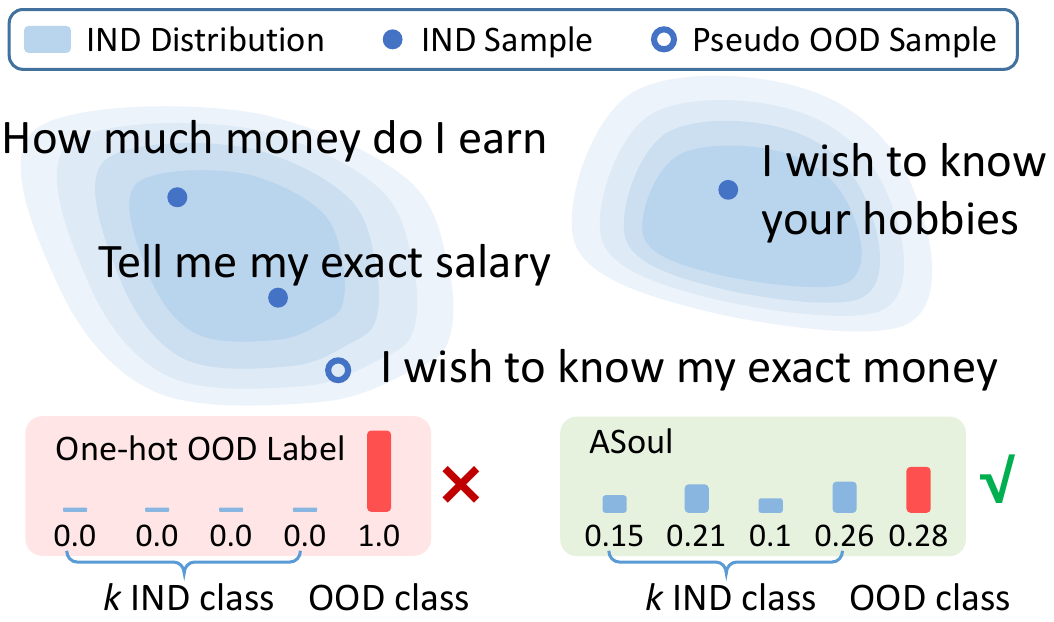}
  \caption{A pseudo OOD sample generated by distorting IND inputs (See more examples in Appendix \ref{append:ood_example}). Comparing to the one-hot OOD label, the soft label produced by ASoul is more suitable for this pseudo OOD sample since it carries some IND intents.
  }
  \label{fig:soft_label}
\end{figure}

Previous studies directly assign one-hot OOD labels to pseudo OOD samples when training the ($k+1$)-way classifier \cite{shu2021odist,chen2021gold}. However, this scheme brings noise to the training process because ``hard'' pseudo OOD samples, i.e., OOD samples that are close to IND distributions, may carry IND intents \cite{zhan2021out} (See Figure \ref{fig:soft_label}). Indiscriminately assigning one-hot OOD labels ignores the semantic connections between pseudo OOD samples and IND intents. Moreover, this issue becomes more severe as most recent studies are dedicated to producing hard pseudo OOD samples \cite{zheng2020out,zhan2021out} since these samples are reported to facilitate OOD detectors better \cite{lee2017training}. Collisions between pseudo OOD samples and IND intents will be more common.

We argue that ideal labels for pseudo OOD samples should be soft labels that allocate non-zero probabilities to all intents \cite{hinton2015distilling,muller2019does}. Specifically, we demonstrate in \secref{sec:pseudo_sample} that pseudo OOD samples generated by most existing approaches should be viewed as unlabeled data since they may carry both IND and OOD intents. Soft labels help capture the semantic connections between pseudo OOD samples and IND intents. Moreover, using soft labels also conforms to the \emph{smoothness assumption}, i.e., samples close to each other are likely to receive similar labels. This assumption lays a foundation for modeling unlabeled data in various previous works \cite{luo2018smooth,van2020survey}.

In this study, we propose an \underline{a}daptive \underline{so}ft pse\underline{u}do \underline{l}abeling (ASoul) method that can estimate soft labels for given pseudo OOD samples and thus help to build better OOD detectors. Specifically, we first construct an embedding graph using supervised contrastive learning to capture semantic connections between pseudo OOD samples and IND intents. Following the smoothness assumption, a graph-smoothed label is produced for each pseudo OOD sample by aggregating nearby nodes on the graph. A co-training framework with two separate classification heads is introduced to refine these graph-smoothed labels. Concretely, the prediction of one head is interpolated with the graph-smoothed label to produce the soft label used to enhance its peer head. The final OOD detector is formulated as a ($k+1$)-way classifier with adaptive decision boundaries.

Extensive experiments on three benchmark datasets demonstrate that ASoul can be used with a wide range of OOD sample generation approaches and consistently improves the OOD detection performance. ASoul also helps achieve new State-of-the-art (SOTA) results on benchmark datasets.
Our major contributions are summarized:

\textbf{1.} We propose ASoul, a method that can estimate soft labels for given pseudo OOD samples. ASoul conforms to the important smoothness assumption for modeling unlabeled data by assigning similar labels to close samples.

\textbf{2.} We construct an embedding graph to help capture the semantic connections between pseudo OOD samples and IND intents. A co-training framework is further introduced to produce the resulting soft labels with the help of two separate classification heads.

\textbf{3.} We conduct extensive experiments on three benchmark datasets. The results show that ASoul consistently improves the OOD detection performance, and it obtains new SOTA results.

\section{Related Work}

\textbf{OOD Detection:}
OOD detection problems have been widely investigated in conventional machine learning studies \cite{geng2020recent}. Recent neural-based methods try to improve the OOD detection performance by learning more robust representations on IND data \cite{zhou-etal-2021-contrastive,zhou2022knn,yan2020unknown,zeng2021adversarial}. These representations can be used to develop density-based or distance-based OOD detectors \cite{lee2018simple,podolskiy2021revisiting,liu2020energy,tan-etal-2019-domain}. Some methods also propose to distinguish OOD inputs using thresholds based methods
\cite{gal2016dropout,lakshminarayanan2017simple,ren2019likelihood,gangal2020likelihood,ryu2017neural}, or utilizing unlabeled IND data \cite{xu-etal-2021-unsupervised,jin2022towards}.

\textbf{Pseudo OOD Sample Generation:} \label{sec:4categorie}
Some works try to tackle OOD detection problems by generating pseudo OOD samples. Generally, four categories of approaches are proposed: 
\textbf{1.} Phrase Distortion \cite{chen2021gold}:
OOD samples are generated by replacing phrases in IND samples;
\textbf{2.} Feature Mixup \cite{zhan2021out}:
OOD features are directly produced by mixing up IND features \cite{zhang2018mixup};
\textbf{3.} Latent Generation \cite{marek2021oodgan}:
OOD samples are drawn from the low-density area of a latent space;
\textbf{4.} Open-domain Sampling \cite{hendrycks2018deep}:
data from other corpora are directly used as pseudo OOD samples.
With these pseudo OOD samples, the OOD detection task can be formalized into a $(k+1)$-way classification problem ($k$ is the number of IND intents). Our method can be combined with all the above OOD generation approaches to improve the OOD detection performance.

\textbf{Soft Labeling:}
Estimating soft labels for inputs has been applied in a wide range of studies such as knowledge distillation \cite{hinton2015distilling,gou2021knowledge,zhang2020dialogue}, confidence calibration \cite{muller2019does,wang2021diversifying}, or domain shift \cite{ng2020ssmba}. However, few studies try to utilize this approach in OOD detection methods. Existing approaches only attempt to assign dynamic weights \cite{ouyang2021energy} or soft labels to IND samples \cite{cheng2022learning}. Our method ASoul is the first attempt to estimate soft labels for pseudo OOD samples.

\textbf{Semi-Supervised Learning:}
Our work is also related to semi-supervised learning (SSL) since they all attempt to utilize unlabeled data and share the same underlying \textit{smoothness assumption} \citep{wang2017theoretical,lee2013pseudo,li2021comatch}. Moreover, the co-training framework in ASoul also helps to enforce the \emph{low-density assumption} (a variant of the smoothness assumption) \citep{van2020survey,chen2022semi} by exploring low-density regions between classes.

\begin{figure}[t] 
\centering 
\includegraphics[scale=0.41]{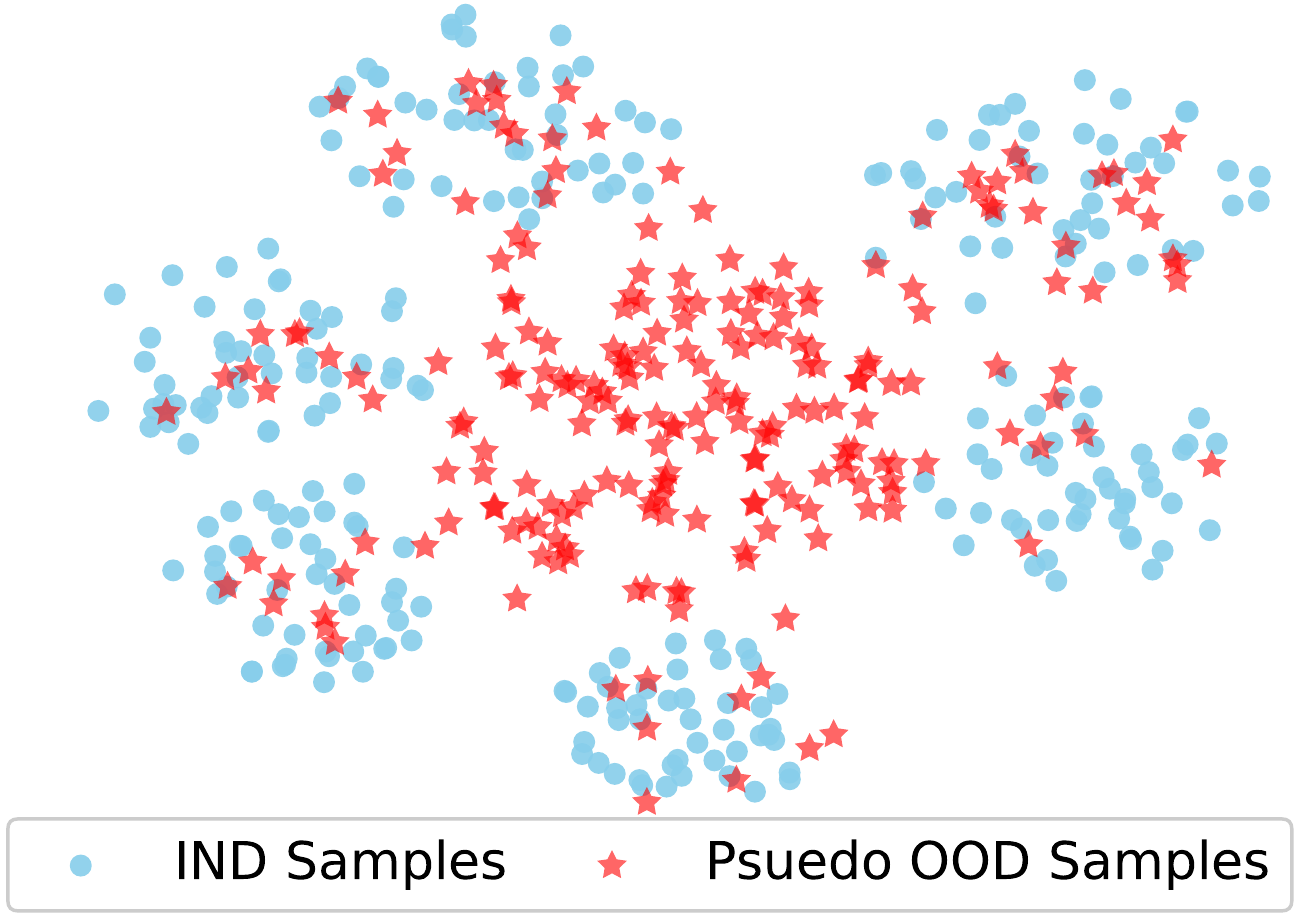} 
\caption{t-SNE visualization of pseudo OOD samples generated by feature mixup \citep{zhan2021out} on the Banking dataset under the 25\% setting. It can be seen that some pseudo OOD samples coincide with IND samples. See more analyses in Appendix \ref{append:ood_example}.}
\label{fig:visulization_mixup} 
\end{figure}

\section{Background}
\subsection{Problem Definition}
OOD intent detectors aim to classify IND intents while detecting OOD inputs. Concretely, given $k$ known IND intent classes $\mathcal{I} = \{I_i\}_{i=1}^k$, the training set $\mathcal{D}_{I}=\{(x_i, y_i)\}$ only contains IND samples, i.e., $x_i$ is an input, and $y_i\in \mathcal{I}$ is the label of $x_i$. The test set $\bar{\mathcal{D}}=\{(\bar{x_i}, \bar{y_i})\}$ consists both IND and OOD samples, i.e., $\bar{y_i} \in \mathcal{I} \cup \{I_{k+1}\}$, in which $I_{k+1}$ is a special OOD intent class. For a testing input $\bar{x_i}$, an OOD detector should classify the intent of $\bar{x_i}$ if $\bar{x_i}$ belongs to an IND intent or reject $\bar{x_i}$ if $\bar{x_i}$ belongs to the OOD intent.

\subsection{Analyzing Pseudo OOD Samples}\label{sec:pseudo_sample}
Recent works have demonstrated that ``hard'' OOD samples, i.e., OOD samples akin to IND distributions, are more efficient in improving the OOD detection performance \cite{lee2018training,zheng2020out}. Promising performances are obtained using these hard samples on various benchmarks \cite{zhan2021out,shu2021odist}.

However, we notice that hard pseudo OOD samples used in previous approaches may coincide with IND samples and carry IND intents. Besides Figure \ref{fig:soft_label}, we further demonstrate this issue by visualizing pseudo OOD samples produced by \citet{zhan2021out}. Specifically, pseudo OOD samples are synthesized using convex combinations of IND features. Figure \ref{fig:visulization_mixup} shows the results on the Banking dataset \cite{casanueva2020efficient} when 25\% intents are randomly selected as IND intents. It can be seen that some pseudo OOD samples fall into the cluster of IND intents, and thus it is improper to assign one-hot OOD labels to these samples. 

The above issue is also observed in other pseudo OOD sample generation approaches. Specifically, we implement the phrase distortion approach proposed by \citet{shu2021odist} and employ crowd-sourced workers to annotate 1,000 generated pseudo OOD samples. Results show that up to 39\% annotated samples carry IND intents (see Appendix \ref{append:ood_example} for more examples).

\section{Method}

\begin{figure}[t] 
\centering 
\includegraphics[scale=0.66]{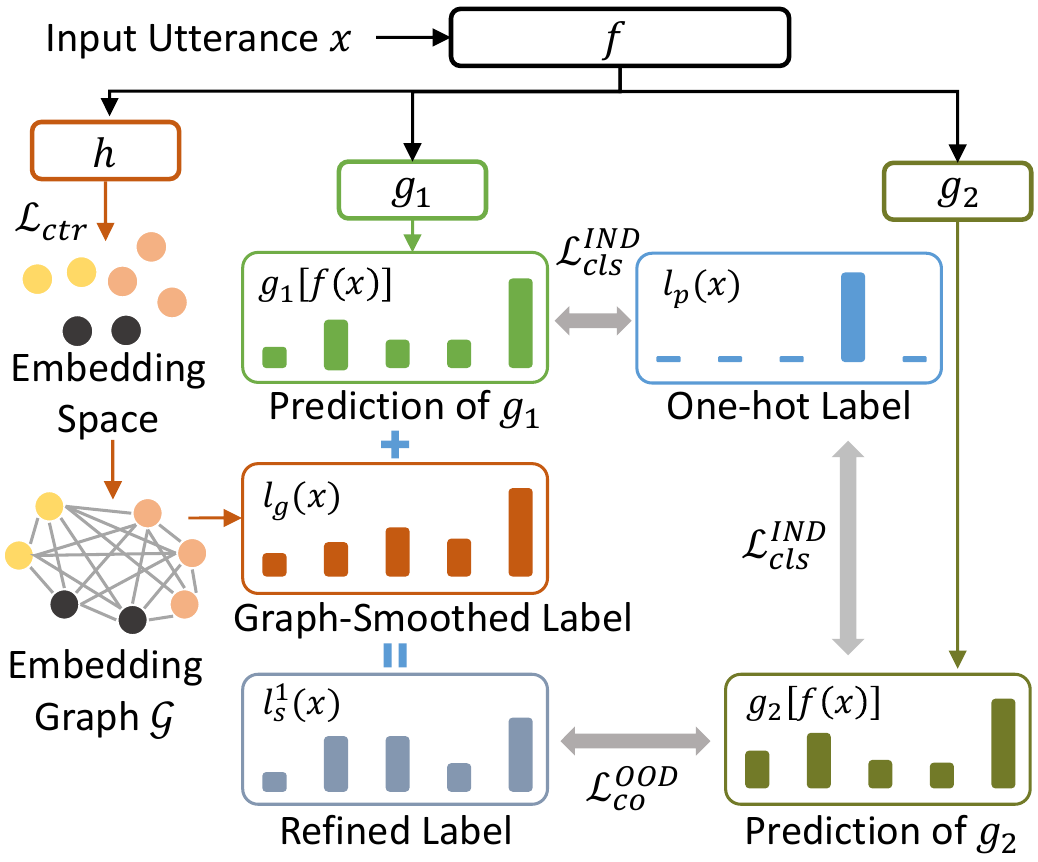} 
\caption{An overview of ASoul. Specifically, an embedding space is obtained using an encoder $f$ and a projection head $h$ by optimizing a supervised constructive loss $\mathcal{L}_{ctr}$ on labeled IND data. A graph-smoothed label $l_g(x)$ conforming to the smoothness assumption is constructed. $l_g(x)$ is further used in a co-training framework, in which two classification heads $g_1$ and $g_2$ are maintained. The prediction of one head is interpolated with $l_g(x)$ to enhance another head.}
\label{fig:framework} 
\end{figure}

\subsection{Overview}\label{sec:overview}
In this study, we build the OOD intent detector following three steps: 1. Construct a set of pseudo OOD samples $\mathcal{D}_P$; 2. Estimate a soft label for each sample $x \in \mathcal{D}_P$; 3. Obtain a $(k+1)$-way classifier and learn a decision boundary for each class to build an OOD detector. A testing input $x$ is identified as OOD if $x$ belongs to the OOD intent $I_{k+1}$ or $x$ is out of all decision boundaries.

Before applying ASoul, we assume a set of pseudo OOD samples $D_P$ are already generated using existing approaches. Figure \ref{fig:framework} shows an overview of ASoul. Specifically, a shared utterance encoder $f$ encodes each input $x \in \mathcal{D}_I \cup \mathcal{D}_P$ into a representation, and an embedding projection head $h$ constructs an embedding graph on these representations. A co-training framework is also implemented using two ($k+1$)-way classification heads $g_1$ and $g_2$, and the prediction of one head is used to enhance soft labels of the peer head.

Note that ASoul is independent of specific methods to produce pseudo OOD samples in $\mathcal{D}_P$. In this study, we test various approaches to obtain $\mathcal{D}_P$.

\subsection{Embedding Graph}

\textbf{Embedding Space:} 
An embedding space is maintained in ASoul to capture semantic of input samples. Specifically, for an input $x_i$, an encoder $f$ is used to convert $x_i$ into a representation vector, then a projection head $h$ is used to map $f(x_i)$ into an L2 normalized embedding $z_i = h[f(x_i)]$ to construct the embedding space. To capture better semantic representations, a supervised contrastive loss \citep{khosla2020supervised,gunel2020supervised} $\mathcal{L}_{ctr}$ is optimized on labeled IND samples in $\mathcal{D}_{I}$:
\begin{equation}\small
\mathcal{L}_{ctr}=\sum_{x_i \in D_I} \frac{-1}{|S(i)|} \sum_{x_j\in S(i)} \log\frac{e^{\Phi(x_i) \cdot \Phi(x_j)/t}}{\sum\limits_{x_k\in A(i)} e^{\Phi(x_i) \cdot \Phi(x_k)/t}},
\end{equation}
in which $S(i)$ represents samples that share the same label with $x_i$ in the current batch, $A(i)$ represents all samples in the current batch except $x_i$, $\Phi$ maps an input $x$ to its corresponding embedding (i.e., $\Phi(x) = h[f(x)]$), and $t > 0$ is a scalar that controls the separation of classes. $\mathcal{L}_{ctr}$ captures the similarities between examples in the same class and contrast them with the examples from different classes \cite{gunel2020supervised}.

\noindent\textbf{Graph-Smoothed Label:}
After obtaining the embedding space, we construct a fully connected unidirectional embedding graph $\mathcal{G}$ using samples in $\mathcal{D}_{IP} = \mathcal{D}_I \cup \mathcal{D}_P$. Specifically, we first map each sample $x \in \mathcal{D}_{IP}$ into an embedding $z$, i.e., $z = \Phi(x)$, and then use all these embeddings as nodes for $\mathcal{G}$. Every two nodes $z_i$ and $z_j$ in $\mathcal{G}$ are linked with an edge. Moreover, we also assign a \emph{prior label} $l_p(x) \in \mathbb{R}^{k+1}$ to each sample $x \in \mathcal{D}_{IP}$ to represent its annotation, i.e., for an IND sample $x \in \mathcal{D}_I$, $l_p(x)$ is defined as the one-hot label corresponding to $y$, and for a pseudo OOD sample $x \in \mathcal{D}_P$, $l_p(x)$ is defined as the one-hot OOD label corresponding to $I_{k+1}$.

For each OOD sample $x \in \mathcal{D}_P$, a graph-smoothed label $l_g(x)$ is obtained by aggregating adjacent nodes on $\mathcal{G}$. Specifically, to conform to the smoothness assumption, we try to minimize the following distance when determining $l_g(x)$:
\begin{equation}
\small
\begin{split}
& \alpha \cdot d[l_g(x), l_p(x)] + (1-\alpha) \sum_{x_j \in \mathcal{D}_{IP}} a_{j} \cdot d[ l_g(x), l_p(x_j) ] \label{eq:lable_obj} \\
& a_{j} = \frac{{\rm exp}(z \cdot z_j / \tau)}{\sum_{k=1}^{|\mathcal{D}_{IP}|} {\rm exp}(z \cdot z_k / \tau)},
\end{split}
\end{equation}
where $0 \leq \alpha \leq 1$ is a scalar, $d$ is a distance function, $\tau > 0$ is a scalar temperature. The second term in Eq. \ref{eq:lable_obj} enforces the smoothness assumption by encouraging $l_g(x)$ to have similar labels with its nearby samples, whereas the first term tries to maintain $l_g(x)$ to meet its original annotation $l_p(x)$. For simplicity, we implement $d$ as the Euclidean distance here, and thus minimizing Eq. \ref{eq:lable_obj} yields:
\begin{equation}\label{eq:graph_smooth}
\small
l_g(x) = \alpha \cdot l_p(x) + (1-\alpha) \sum_{x_j \in \mathcal{D}_{IP}} a_{j} \cdot l_p(x_j)
\end{equation}

Note that the result we derived in Eq. \ref{eq:graph_smooth} follows most previous graph-smoothing approaches in semi-supervised learning \cite{van2020survey}. To the best of our knowledge, we are the first to apply this scheme to OOD detection tasks.

\subsection{Co-Training Framework}
To further enforce the smoothness assumption, a co-training framework is introduced in ASoul to learn better soft labels using $l_g(x)$. Specifically, we implement two classification heads $g_1$ and $g_2$ on top of the shared encoder $f$. Each classification head $g_i$ maps the output of $f$ to a ($k+1$) dimensional distribution, i.e, $g_i[f(x)] \in \mathbb{R}^{k+1}$ $(i=1, 2)$, and a classification loss is optimized on IND samples:
\begin{equation}\label{eq:cls_ind}
\small
\mathcal{L}_{cls}^{IND} = \sum_{x \in \mathcal{D}_{I}} \frac{1}{2} \sum_{i=1}^2 CE(l_p(x) , g_i[f(x)]),
\end{equation}
in which $CE$ measures the cross-entropy between two distributions.

Besides optimizing $\mathcal{L}_{cls}^{IND}$, a co-training process is implemented to refine $l_g(x)$ for each $x \in \mathcal{D}_P$. Specifically, a soft label $l_s^1(x)$ (or $l_s^2(x)$) is produced by interpolating $l_g(x)$ with the prediction of one classification head $g_1$ (or $g_2$), and the resulting soft label is used to optimize another head $g_2$ (or $g_1$). Concretely, the following co-training loss is optimized:
\begin{equation}\label{eq:co_ood}
\small
\begin{aligned}
\mathcal{L}_{co}^{OOD} &= \sum_{x \in \mathcal{D}_{P}} \frac{1}{2} \sum_{j=1}^2 \sum_{i=1}^{2} \mathbf{1}_{i \neq j} CE( l_s^i(x)  , g_j[f(x)]), \\
l_s^i(x) &= \beta \cdot l_g(x) + (1-\beta) \cdot g_i[f(x)], (i=1,2),
\end{aligned}
\end{equation}
where $0 \leq \beta \leq 1$ is a weight scalar. Different dropout masks are used in $g_1$ and $g_2$ to prompt the diversity required by co-training. Note that as indicated by \citet{lee2013pseudo} and \citet{chen2022semi}, the co-training loss $\mathcal{L}_{co}^{OOD}$ favors low density separation between classes, and thus it helps to enforce the low-density assumption when training $g_i$.

The overall training loss for our method is:
\begin{equation}\label{eq:total_loss}
\mathcal{L}=\mathcal{L}_{ctr} + \mathcal{L}_{cls}^{IND} + \mathcal{L}_{co}^{OOD}
\end{equation}

\subsection{OOD Detection}\label{sec:adb}
In the inference phase, we directly use the averaged prediction of $g_1$ and $g_2$ to implement the OOD detector $g(y | x) \in \mathbb{R}^{k+1}$.
\begin{equation}
\small
g(y | x) = (g_1[f(x)] + g_2[f(x)]) / 2
\end{equation}
Moreover, an adaptive decision boundary (ADB) is learnt on top of $g(y|x)$ to further reduce the open space risk \citep{zhou2022knn,shu2021odist}.
Specifically, we follow the approach of \citet{zhang2021deep} to obtain a central vector $c_i$ and a decision boundary scalar $b_i$ for each intent class $I_i \in \mathcal{I} \cup \{I_{k+1}\}$.
In the testing phase, the label $y$ for each input $x$ is obtained as:
\begin{equation*}
\small
y=\left\{
\begin{aligned}
& I_{k+1}, ~~~ {\rm if} ~~ || f(x) - c_i|| > b_i,\forall{i}\in \{1, \dots ,k+1\},\\
& \argmax_{I_j\in \mathcal{I} \cup \{I_{k+1}\}} g(I_j | x), ~~~ {\rm otherwise},
\end{aligned}
\right.
\end{equation*}
In this way, we can classify IND intents while rejecting OOD intent.

\section{Experiments}
\subsection{Datasets}
Following most previous works \citep{zhang2021deep,shu2021odist}, we use three benchmark datasets: \textbf{CLINC150} \citep{larson2019evaluation} contains 150 IND intents and one OOD intent. We follow \citet{zhan2021out} to group all samples from the OOD intent into the test set; \textbf{StackOverflow} \citep{xu2015short} contains 20 classes with 1,000 samples in each class; \textbf{Banking} \citep{casanueva2020efficient} contains 77 intents in the banking domain. Standard splits of above datasets is followed (See Table \ref{tab:datasets}).

\begin{table}
\begin{center}
\small
\begin{tabular}{  l|c|c|c|c  } 
\toprule
 Dataset & Train & Valid & Test & \#Intent \\ 
\midrule
 CLINC150 & 15,000 & 3,000 & 5,700 & 150 \\ 
 StackOverflow & 12,000 & 2,000 & 6,000 & 20 \\ 
 Banking & 9,003 & 1,000 & 3,080 & 77 \\
\bottomrule
\end{tabular}
\end{center}
\caption{Dataset statistics.}
\label{tab:datasets}
\end{table}

\subsection{Implementation Details}
Our encoder $f$ is implemented using BERT \cite{devlin2018bert} with a mean-pooling layer. The projection head $h$, classification heads $g1$ and $g_2$ are implemented as two-layer MLPs with the LeakyReLU activation \cite{xu2020reluplex}. The optimizer AdamW and Adam \cite{kingma2014adam} is used to finetune BERT and all the heads with a learning rate of 1e-5 and 1e-4, respectively. We use $\tau=0.1$, $\alpha=0.11$ and $\beta=0.9$ in all experiments. All results reported in our paper are averages of 10 runs with different random seeds. See Appendix \ref{append:implementation_details} for more implementation details.
Note that ASoul only introduces little computational overhead compared to the vanilla BERT model (See Appendix \ref{append:comp}.),
and we detail how to choose important hyper-parameters for ASoul in Appendix \ref{append:hyper_sen}.

\begin{table*}[t]
\small
\setlength\tabcolsep{1.5pt} 
\begin{tabular}{cl|cccc|cccc|cccc}
\toprule
\multirow{2}{*}{}  &  &\multicolumn{4}{c|}{CLINC150}  &  \multicolumn{4}{c|}{StackOverflow}  &\multicolumn{4}{c}{Banking}     \\
& Methods  & Acc-All & F1-All & F1-OOD & F1-IND &  Acc-All &   F1-All & F1-OOD & F1-IND & Acc-All &  F1-All & F1-OOD & F1-IND \\
\midrule
\multirow{8}{*}{25\%} 
  & LG+Onehot  & 30.54      & 39.16   & 24.25 & 39.55   & 42.45   & 43.20  & 44.43  & 42.95  & 26.72    & 48.04  & 9.03  & 50.09    \\
  & OS+Onehot  & 45.56      & 47.04   & 48.81 & 47.00   & 26.93   & 38.28  & 8.89   & 44.16  & 29.86    & 44.09  & 14.92 & 45.62   \\
  & PD+Onehot  & 89.79      & 80.04   & 93.42 & 79.69   & 91.53   & 85.05  & 94.41  & 83.18  & 81.69    & 73.44  & 87.11 & 72.72   \\
  & FM+Onehot  & 89.78      & 82.04   & 93.34 & 81.74   & 89.76   & 76.95  & 93.60  & 73.62  & 79.42    & 72.70  & 84.87 & 72.06    \\
\cmidrule{2-14}
  & LG+ASoul & 32.38          & 40.70  & 27.45 & 41.05    & 58.00          & 53.97          & 64.81          & 51.80          & 29.06          & 48.81       & 13.83 & 50.66   \\
  & OS+ASoul & 50.67          & 49.70  & 55.80 & 49.54    & 27.23          & 41.29          & 9.66           & 47.61          & 33.38          & 47.34     & 22.14 & 48.67     \\
  & PD+ASoul & 90.61          & 81.68  & 93.95 & 81.36    & \textbf{93.20} & \textbf{87.01} & \textbf{95.58} & \textbf{85.30} & 85.32          & \textbf{78.64} & 89.80 & \textbf{78.06}\\
  & FM+ASoul & \textbf{92.71} & \textbf{84.11} & \textbf{95.42} & \textbf{83.81} & 92.04          & 86.03   & 94.76 & 84.29   & \textbf{87.41} & 78.39     & \textbf{91.52} & 77.70     \\
\midrule
\midrule
\multirow{8}{*}{50\%} 
  & LG+Onehot & 42.51 & 61.71 & 12.89 & 62.36 & 60.00 & 67.42 & 52.92 & 68.87 & 47.89 & 66.84 & 4.37 & 68.49 \\
  & OS+Onehot & 55.44 & 66.88 & 43.80  & 67.19 & 47.92 & 60.26 & 7.93 & 65.49 & 49.98 & 66.82 & 11.82 & 68.26\\
  & PD+Onehot & 88.61 & 86.57 & 90.62 & 86.52 & 88.52 & 87.35 & 89.57 & 87.13 & 80.90 & 81.78 & 81.32 & 81.79\\
  & FM+Onehot & 87.91 & 87.06 & 89.71 & 87.03 & 83.47 & 80.78 & 85.48 & 80.31 & 80.32 & 81.48 & 80.57 &  81.50 \\
\cmidrule{2-14}
  & LG+ASoul & 46.53 & 63.09 & 24.26 & 63.60 & 63.25 & 71.54 & 58.03 & 72.90 & 50.45 & 68.10 & 12.63 & 69.56\\
  & OS+ASoul & 58.19 & 68.46 & 49.04 & 68.72 & 49.13 & 62.52 & 12.11 & 67.56 & 50.61 & 68.52 & 12.33 & 70.00\\
  & PD+ASoul & 89.50 & 87.54 & 91.40 & 87.49 & \textbf{89.45} & \textbf{88.65} & \textbf{90.46} & \textbf{88.47} & 81.86 & 83.84 & 81.51 & 83.90\\
  & FM+ASoul & \textbf{89.96} & \textbf{88.20} & \textbf{91.72} & \textbf{88.15} & 88.92 & 88.24 & 89.69 & 88.09 & \textbf{81.98} & \textbf{83.96} & \textbf{81.65} & \textbf{84.03}\\
\midrule
\midrule
\multirow{8}{*}{75\%} 
  & LG+Onehot & 57.82 & 74.64 & 7.47 & 75.24 & 69.58 & 76.84 & 16.52 & 80.86 & 72.31 & 82.34 & 13.32 & 83.53\\
  & OS+Onehot & 69.54 & 80.00 & 47.04 & 80.29 & 68.81 & 75.33 & 5.39 & 80.00 & 71.48 & 81.66 & 7.97 & 82.92\\
  & PD+Onehot & 87.70 & 89.30 & 85.86 & 89.33 & 83.75 & 86.88 & 75.21 & 87.66 & 82.79 & 86.94 & 71.95 & 87.20\\
  & FM+Onehot & 88.59 & 89.66 & 87.18 & 89.68 & 83.44 & 86.67 & 73.62 & 87.53 & 82.06 & 87.31 & 65.60 & 87.68 \\
\cmidrule{2-14}
  & LG+ASoul & 59.70 & 75.62 & 13.06 & 76.19 & 71.07 & 77.76 & 20.09 & 81.61 & 73.64 & 83.51 & 19.58 & 84.61\\
  & OS+ASoul & 71.60 & 81.25 & 52.20 & 81.51 & 69.51 & 76.67 & 8.24 & 81.23 & 72.23 & 82.44 & 10.97 & 83.68\\
  & PD+ASoul & 88.59 & 90.57 & 86.45 & 90.60 & 84.53 & 87.44 & 73.75 & 88.35 & 83.30 & 87.76 & 69.03 & 88.09 \\
  & FM+ASoul & \textbf{89.88} & \textbf{91.38} & \textbf{88.21} & \textbf{91.41} & \textbf{85.00} & \textbf{87.90} & \textbf{75.76} & \textbf{88.71} & \textbf{84.47} & \textbf{88.39} & \textbf{72.64} & \textbf{88.66} \\
\bottomrule
\end{tabular}
\centering
\caption{
Performances of ASoul when combined with different OOD sample generation approaches. 
Best results among each setting are bolded.
The best performing ASoul-based method significantly outperforms other baselines with $p$-value < 0.05 ($t$-test) in each setting.}
\label{tab:res_onehot_soft}
\end{table*}

\subsection{Experiment Setups and Baselines}\label{sec:baselines}
Following \cite{zhang2021deep,zhan2021out,shu2021odist}, we randomly sample 25\%, 50\%, and 75\% intents as the IND intents and regard all remaining intents as one OOD intent $I_{k+1}$. Note that in the training and validation process, we only use samples from the IND intents. Hyper-parameters are searched based on IND intent classification performances on validation sets.

To validate our claim that ASoul is independent of specific methods to produce $\mathcal{D}_P$. 
We tested the performance of ASoul with four pseudo OOD sample generation approaches:
1. \emph{Phrase Distortion} (\textbf{PD}): follows \citet{shu2021odist} to generates OOD samples by distorting IND samples;
2. \emph{Feature Mixup} (\textbf{FM}): follows \citet{zhan2021out} to produce OOD features using convex combinations of IND features;
3. \emph{Latent Generation} (\textbf{LG}): follows \citet{zheng2020out} to decode pseudo OOD samples from a latent space;
4. \emph{Open-domain Sampling} (\textbf{OS}): follows \citet{zhan2021out} to use sentences from other corpora as OOD samples. 
Each approach mentioned above associates with one of the four categories listed in \secref{sec:4categorie}.

Moreover, we also applied the above pseudo OOD sample generation approaches with the previous SOTA method that uses one-hot labeled pseudo OOD samples \cite{shu2021odist}.
Specifically, a ($k+1$)-way classifier is trained by optimizing the cross-entropy loss on $\mathcal{D}_I \cup \mathcal{D}_P$ using one-hot labels,
and the ADB approach presented in \secref{sec:adb} is used to construct the OOD detector.

We also compared our method to other competitive OOD detection baselines:
\textbf{MSP:} \citep{hendrycks2017baseline} utilizes the maximum Softmax probability of a $k$-way classifier to detect OOD inputs;
\textbf{DOC:} \citep{shu2017doc} employs $k$ 1-vs-rest Sigmoid classifiers and use the maximum predictions to detect OOD intents;
\textbf{OpenMax:} \citep{bendale2016towards} fits a Weibull distribution to the logits and re-calibrates the confidences with an Open-Max Layer;
\textbf{LMCL:} \citep{lin2019deep} introduces a large margin cosine loss to maximize the decision margin and uses LOF as the OOD detector;
\textbf{ADB:} \citep{zhang2021deep} learns an adaptive decision boundaries for OOD detection;
\textbf{Outlier:} \citep{zhan2021out} mixes convex interpolated outliers and open-domain outliers to train a ($k+1$)-way classifier;
\textbf{SCL:} \citep{zeng-etal-2021-modeling} uses a supervised contrastive learning loss to separate IND and OOD features;
\textbf{GOT:} \citep{ouyang2021energy} shapes an energy gap between IND and OOD samples.
\textbf{ODIST:} \citep{shu2021odist} generates pseudo OOD samples with using a pre-trained language model.

For fair comparisons, all baselines are implemented with codes released by their authors, and use BERT as the backbone. For threshold-based baselines, 100 OOD samples are used in the validation to determine the thresholds used for testing. See Appendix \ref{append:baseline} for more details about baselines.

\begin{table*}[t]
\small
\setlength\tabcolsep{2pt} 
\begin{tabular}{cl|cccc|cccc|cccc}
\toprule
\multirow{2}{*}{}  &  &\multicolumn{4}{c|}{CLINC150}  &  \multicolumn{4}{c|}{StackOverflow}  &\multicolumn{4}{c}{Banking} \\
& Methods  & Acc-All & F1-All & F1-OOD & F1-IND &  Acc-All &   F1-All & F1-OOD & F1-IND & Acc-All &  F1-All & F1-OOD & F1-IND\\
\midrule
\multirow{10}{*}{25\%} 
    & MSP     & 47.02      & 47.62  & 50.88 & 47.53    & 28.67      & 37.85   & 13.03 & 42.82   & 43.67      & 50.09  & 41.43 & 50.55    \\
    & DOC     & 74.97      & 66.37  & 81.98 & 65.96   & 42.74      & 47.73  & 41.25 & 49.02    & 56.99      & 58.03   & 61.42 & 57.85   \\
    & OpenMax & 68.50      & 61.99  & 75.76 & 61.62    & 40.28      & 45.98  & 36.41 & 47.89    & 49.94      & 54.14    & 51.32 & 54.28  \\
    & SCL     & 75.01      & 65.45  & 81.92 & 65.01    & 62.08      & 61.01  & 67.99 & 59.61    & 70.82      & 64.82   & 77.28 & 64.17   \\
    & GOT     & 72.63      & 64.01  & 79.45 & 63.60   & 65.02      & 62.26   & 68.58 & 61.00   & 63.05      & 63.49    & 68.61 & 63.22  \\
    & LMCL    & 81.43      & 71.16  & 87.33 & 70.73    & 47.84      & 52.05  & 49.29 & 52.60    & 64.21      & 61.36  & 70.44 & 60.88    \\
    & ADB     & 87.59      & 77.19  & 91.84 & 76.80    & 86.72      & 80.83  & 90.88 & 78.82    & 78.85      & 71.62  & 84.56 & 70.94    \\
    & Outlier & 88.44      & 80.73  & 92.35 & 80.43    & 68.74      & 65.64  & 74.86 & 63.80    & 74.11      & 69.93   & 80.12 & 69.39    \\
    & ODIST   & 89.79      & 80.04  & 93.42 & 79.69   & 91.53      & 85.05   & 94.41 & 83.18   & 81.69      & 73.44   & 87.11 & 72.72   \\
\cmidrule{2-14}
    & FM+ASoul & \textbf{92.71} & \textbf{84.11} & \textbf{95.42} & \textbf{83.81} & \textbf{92.04}          & \textbf{86.03}   & \textbf{94.76} & \textbf{84.29}         & \textbf{87.41} & \textbf{78.39}     & \textbf{91.52} & \textbf{77.70}     \\
\midrule
\midrule
\multirow{10}{*}{50\%} 
    & MSP & 62.96 & 70.41 & 57.62 & 70.58 & 52.42 & 63.01 & 23.99 & 66.91 & 59.73 & 71.18 & 41.19 & 71.97\\
    & DOC & 77.16 & 78.26 & 79.00 & 78.25 & 52.53 & 62.84 & 25.44 & 66.58 & 64.81 & 73.12 & 55.14 & 73.59 \\
    & OpenMax & 80.11 & 80.56 & 81.89 & 80.54 & 60.35 & 68.18 & 45.00 & 70.49 & 65.31 & 74.24 & 54.33 & 74.76 \\
    & SCL & 71.14 & 75.03 & 70.81 & 75.09 & 76.16 & 78.95 & 74.42 & 79.40 & 74.81 & 78.04 & 72.45 & 78.19 \\
    & GOT & 67.06 & 73.15 & 63.48 & 73.28 & 65.56 & 72.19 & 55.53 & 73.86 & 69.97 & 76.37 & 63.03 & 76.72 \\
    & LMCL & 83.35 & 82.16 & 85.85 & 82.11 & 58.98 & 68.01 & 43.01 & 70.51 & 72.73 & 77.53 & 69.53 & 77.74 \\
    & ADB & 86.54 & 85.05 & 88.65 & 85.00 & 86.40 & 85.83 & 87.34 & 85.68 & 78.86 & 80.90 & 78.44 & 80.96 \\
    & Outlier & 88.33 & 86.67 & 90.30 & 86.54 & 75.08 & 78.55 & 71.88 & 79.22 & 72.69 & 79.21 & 67.26 & 79.52 \\
    & ODIST & 88.61 & 86.57 & 90.62 & 86.52 & 88.52 & 87.35 & 89.57 & 87.13 & 80.90 & 81.78 & 81.32 & 81.79\\
\cmidrule{2-14}
    & FM+ASoul & \textbf{89.96} & \textbf{88.20} & \textbf{91.72} & \textbf{88.15} & \textbf{88.92} & \textbf{88.24} & \textbf{89.69} & \textbf{88.09} & \textbf{81.98} & \textbf{83.96} & \textbf{81.65} & \textbf{84.03}\\
\midrule
\midrule
\multirow{10}{*}{75\%} 
    & MSP & 74.07 & 82.38 & 59.08 & 82.59 & 72.17 & 77.95 & 33.96 & 80.88 & 75.89 & 83.60 & 39.23 & 84.36\\
    & DOC & 78.73 & 83.59 & 72.87 & 83.69 & 68.91 & 75.06 & 16.76 & 78.95 & 76.77 & 83.34 & 50.60 & 83.91 \\
    & OpenMax & 76.80 & 73.16 & 76.35 & 73.13 & 74.42 & 79.78 & 44.87 & 82.11 & 77.45 & 84.07 & 50.85 & 84.64 \\
    & SCL & 76.50 & 82.65 & 66.90 & 82.79 & 79.91 & 84.41 & 63.79 & 85.78 & 78.45 & 84.09 & 56.19 & 84.57 \\
    & GOT & 72.65 & 81.49 & 54.11 & 81.73 & 77.76 & 81.85 & 52.80 & 83.79 & 77.11 & 83.36 & 48.30 & 83.97\\
    & LMCL & 83.71 & 86.23 & 81.15 & 86.27 & 72.33 & 78.28 & 37.59 & 81.00 & 78.52 & 84.31 & 58.54 & 84.75 \\
    & ADB & 86.32  & 88.53 & 83.92 & 88.58  & 82.78 & 85.99 & 73.86 & 86.80 & 81.08 & 85.96 & 66.47 & 86.29\\
    & Outlier & 88.08 & 89.43 & 86.28 & 89.46 & 81.71 & 85.85 & 65.44 & 87.22 & 81.07 & 86.98 & 60.71 & 87.47\\
    & ODIST & 87.70 & 89.30 & 85.86 & 89.33 & 83.75 & 86.88 & 75.21 & 87.66 & 82.79 & 86.94 & 71.95 & 87.20\\
\cmidrule{2-14}
    & FM+ASoul & \textbf{89.88} & \textbf{91.38} & \textbf{88.21} & \textbf{91.41} & \textbf{85.00} & \textbf{87.90} & \textbf{75.76} & \textbf{88.71} & \textbf{84.47} & \textbf{88.39} & \textbf{72.64} & \textbf{88.66} \\
\bottomrule
\end{tabular}
\centering
\caption{Performance of ASoul and baselines. Best results among each setting are bolded. All improvements of our method over baselines are significant with $p$-value < 0.05 ($t$-test).}
\label{tab:baselines}
\end{table*}

\subsection{Metrics}
Following \citet{zhang2021deep,zhan2021out,shu2021odist}, we use overall accuracy (\textbf{Acc-All}) and macro F1-score (\textbf{F1-All}) calculated over all intents (IND and OOD intents) to evaluate the OOD detection performance. We also calculate macro F1-scores over IND intents (\textbf{F1-IND}) and OOD intent (\textbf{F1-OOD}) to evaluate fine-grained performances.

\subsection{Results}
Table~\ref{tab:res_onehot_soft} shows the OOD detection performance associated with different pseudo OOD sample generation approaches.
Specifically, results marked with ``ASoul'' measures the performance of our method, 
while results marked with ``Onehot'' correspond to the performance of the previous SOTA method \cite{shu2021odist} that uses one-hot labeled samples.
We can observe that:
\textbf{1.} ASoul consistently outperforms its one-hot labeled counterpart with large margins.
This validates our claim that ASoul can be used to improve the OOD detection performance with different pseudo OOD sample generation approaches;
\textbf{2.} ``hard'' pseudo OOD samples yield by \textbf{FM} lead to sub-optimal performance when assigned with one-hot labels (i.e., FM+Onehot generally under-performs PD+Onehot),
while it achieves the best performance when combined with ASoul.
This demonstrates that assigning one-hot labels to hard pseudo OOD samples introduces noise to the training process and ASoul helps to alleviate these noises. 
\textbf{3.} Although OOD samples yielded by the open-domain sampling approach are usually disjoint from the training task, they still benefit from ASoul.
We suppose this is because the soft labels produced by ASoul prevent the OOD detector from becoming over-confident, which is important to improve the OOD detection performance.

Table~\ref{tab:baselines} shows the performance of all baselines and our best method FM+ASoul.
It can be seen that FM+ASoul significantly outperforms all baselines and achieve SOTA results on all three datasets.
This validates the effectiveness of ASoul in improving the OOD detection performance.
We can also observe large improvements of ASoul when labeled IND datasets are small  (i.e., in 25\% and 50\% settings).
This demonstrates the potential of ASoul to be applied in practical scenarios, 
particularly in the early phases of the development that we usually need to handle a large number of OOD inputs with limited IND intents \cite{zhan2021out}.

\begin{table*}[t]
\small
\setlength\tabcolsep{1.0pt} 
\begin{tabular}{l|cccc|cccc|cccc}
\toprule
\multirow{2}{*}{Methods}    & \multicolumn{4}{c|}{25\%} & \multicolumn{4}{c|}{50\%} & \multicolumn{4}{c}{75\%} \\
                            & Acc-All & F1-ALL & F1-OOD & F1-IND & Acc-All & F1-ALL  & F1-OOD & F1-IND  & Acc-All & F1-ALL  & F1-OOD & F1-IND  \\
\midrule
PD+ASoul & \textbf{90.61} & \textbf{81.68} & \textbf{93.95} & \textbf{81.36} & \textbf{89.50}  & \textbf{87.54} & \textbf{91.40} & \textbf{87.49} & \textbf{88.59} & \textbf{90.57} & \textbf{86.45} & \textbf{90.60} \\
PD+ASoul-CT & 90.42 & 81.36 & 93.82 & 81.04 & 88.36 & 87.05 & 90.20 & 87.01 & 88.28 & 90.14 & 86.16 & 90.18 \\
PD+ASoul-GS & 89.96 & 80.66 & 93.55 & 80.32 & 87.96 & 86.93 & 89.81 & 86.89 & 88.04 & 90.03 & 85.92 & 90.06 \\
PD+USoul & 89.33 & 80.20 & 93.05 & 79.87 & 87.72  & 86.59 & 89.60 & 86.55 & 87.21 & 89.29 & 85.02 & 89.33 \\
PD+KnowD & 89.49 & 79.46 & 93.33 & 79.10 & 87.09 & 85.51 & 89.05 & 85.47 & 85.79 & 88.11 & 83.20 & 88.16\\
\midrule
FM+ASoul & \textbf{92.71} & \textbf{84.11} & \textbf{95.42} & \textbf{83.81} & \textbf{89.96} & \textbf{88.20} & \textbf{91.72} & \textbf{88.15} & \textbf{89.88} & \textbf{91.38} & \textbf{88.21} & \textbf{91.41} \\
FM+ASoul-CT & 92.47 & 83.43 & 95.28 & 83.11 &88.58 & 87.50 & 90.28 & 87.46 & 89.02& 90.76 & 87.09 & 90.80 \\
FM+ASoul-GS & 91.73 & 82.92 & 94.77 & 82.61 &88.98 & 87.31 & 90.73 & 87.27 &88.82 & 90.50 & 86.96 & 90.53 \\
FM+USoul & 90.35 & 82.13 & 93.80 & 81.82 &87.51& 87.02 & 89.17 & 86.99 &88.21& 89.70 & 86.46 & 89.73 \\
FM+KnowD & 90.11 & 81.84 & 93.58 & 81.53 & 88.18 & 86.23 & 90.17 & 86.18 & 86.54 & 88.64 & 84.31 & 88.68 \\
\bottomrule
\end{tabular}
\centering
\caption{Ablation study results on the CLINC150 dataset.}
\label{tab:ablation}
\end{table*}

\subsection{Ablation Study}
Ablation studies were performed to verify the effect of each component in ASoul:
We tested following variants:
1. \textbf{ASoul-CT} removes the co-training framework, i.e., only one classification head $g_1$ is implemented without the co-training process.
In this variant, the loss shown in Eq.\ref{eq:total_loss} is optimized by moving $g_2$ and setting $\beta=1$ in Eq.\ref{eq:co_ood}.
2. \textbf{ASoul-GS} removes the graph-smoothed labels, i.e., the embedding graph is not constructed. 
In this variant, losses shown in Eq.\ref{eq:cls_ind} and \ref{eq:co_ood} are optimized and $l_g(x)$ in Eq.\ref{eq:co_ood} is replaced with the one-hot prior label $l_p(x)$.
3. \textbf{USoul} employs uniformly distributed soft labels for samples in $\mathcal{D}_P$.
In this variant, the soft label $l_s^i(x)$ in Eq.\ref{eq:co_ood} is obtained by uniformly reallocating a small portion of OOD probability to OOD intents.
4. \textbf{KnowD} implements a knowledge distillation process to obtain soft labels,
i.e., a $k$-way IND intent classifier is first trained on $\mathcal{D}_I$ and its predictions are interpolated with the one-hot OOD label to obtain the soft label $l_s^i(x)$ in Eq.\ref{eq:co_ood}.

All above variants are tested with two approaches to produce $\mathcal{D}_P$: PD and FM. Results in Table \ref{tab:ablation} indicate that our method outperforms all ablation models.
We can further observe that:
\textbf{1.} soft-labels obtained using other approaches degenerate the model performance by a large margin.
This shows the effectiveness of the soft labels produced by ASoul.
\textbf{2.} graph-smoothed labels bring the largest improvement compared to other components.
This further proves the importance of modeling semantic connections between OOD samples and IND intents.

\subsection{Feature Visualization}
\begin{figure}[t] 
\centering 
\includegraphics[scale=0.40]{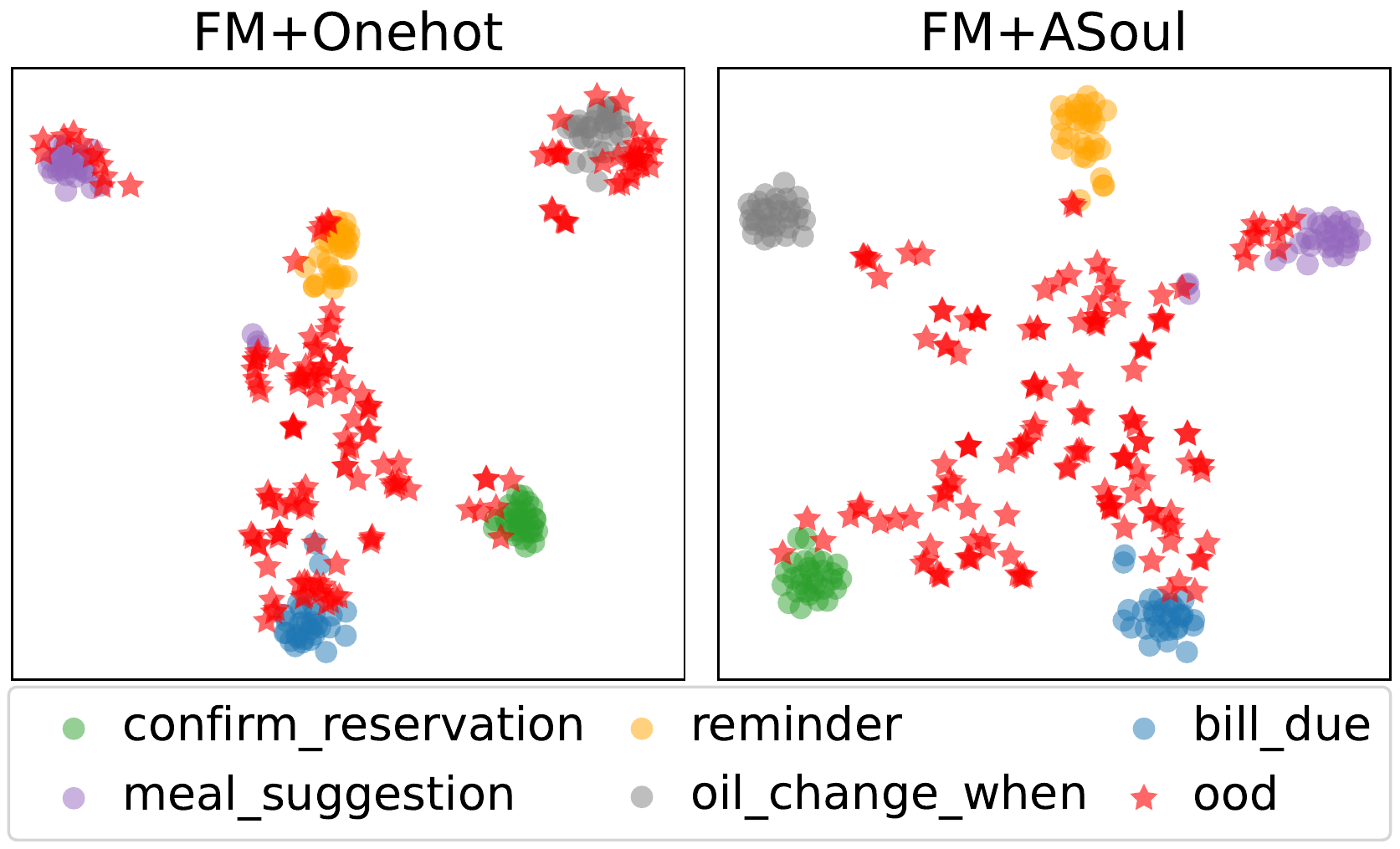} 
\caption{t-SNE visualization of learnt features on the test set of CLINC150 under the 25\% setting.} 
\label{fig:visualize} 
\end{figure}

To further demonstrate the effectiveness of ASoul, we visualized the features learnt in the penultimate layer of OOD detectors that are trained using one-hot labels or soft labels. We use the best performing pseudo OOD samples generation approach (i.e., FM) in this analysis. Results shown in Figure \ref{fig:visualize} demonstrate that soft labels produced by ASoul help the OOD detector learn better representations compared to one-hot labels. The learnt feature space is smoother and representations for IND and OOD samples are more separable. This validates our claim that ASoul helps to conform to the smoothness assumption and improves the OOD detection performance.

\section{Conclusion}
In this paper, we first analyze the limitation of existing OOD detection approaches that use one-hot labeled pseudo OOD samples. Then we propose a method ASoul that can estimate soft labels for given pseudo OOD samples and use these soft labels to train better OOD detectors. An embedding space is constructed to produce graph-smoothed labels to capture the semantic connections between OOD samples and IND intents. A co-training framework further refines these graph-smoothed labels. Experiments demonstrate that our method can be combined with different pseudo OOD sample generation approaches, and it helps achieve SOTA results on three benchmark datasets.
In the future, we plan to apply our method in other tasks, such as Text-to-SQL parsers \citep{hui2021dynamic,wang2022proton, qin2022sun} or lifelong learning \citep{dai2022lifelong}.

\section*{Limitations}
We identify the major limitation of this work is its input modality. Specifically, our method is limited to textual inputs and ignores inputs in other modalities such as vision, audio, or robotic features. These modalities provide valuable information that can be used to better OOD detectors. Fortunately, with the help of multi-modal pre-training models \cite{radford2021learning,zheng-etal-2022-mmchat}, we can obtain robust features well aligned across different modalities. In future works, we will try to model multi-modal contexts for OOD detection and explore better pseudo OOD sample generation approaches.

Another limitation of this work is the pre-training model used in experiments: a model pre-trained on dialogue corpora is expected to yield better performance \cite{he2022galaxy,he2022space,he2022unified,zhou2021eva,wang2020large,zheng2020pre}. Moreover, it is reported that better OOD detection performance can be obtained if we can extract more robust features for IND tasks \cite{vaze2021open}. Our method can be readily applied to other feature extractors that are better performed on dialogues.

\section*{Ethics Statement}
This work does not present any direct ethical issues. In the proposed work, we seek to develop a general method for OOD intent detection, and we believe this study leads to intellectual merits that benefit from a reliable application of NLU models.  All experiments are conducted on open datasets.

\bibliography{anthology,custom}
\bibliographystyle{acl_natbib}

\appendix

\section{More Examples of Pseudo OOD Samples}\label{append:ood_example}
This appendix shows more pseudo OOD samples that are generated using existing approaches.

\begin{figure}[t] 
\centering 
\includegraphics[scale=0.41]{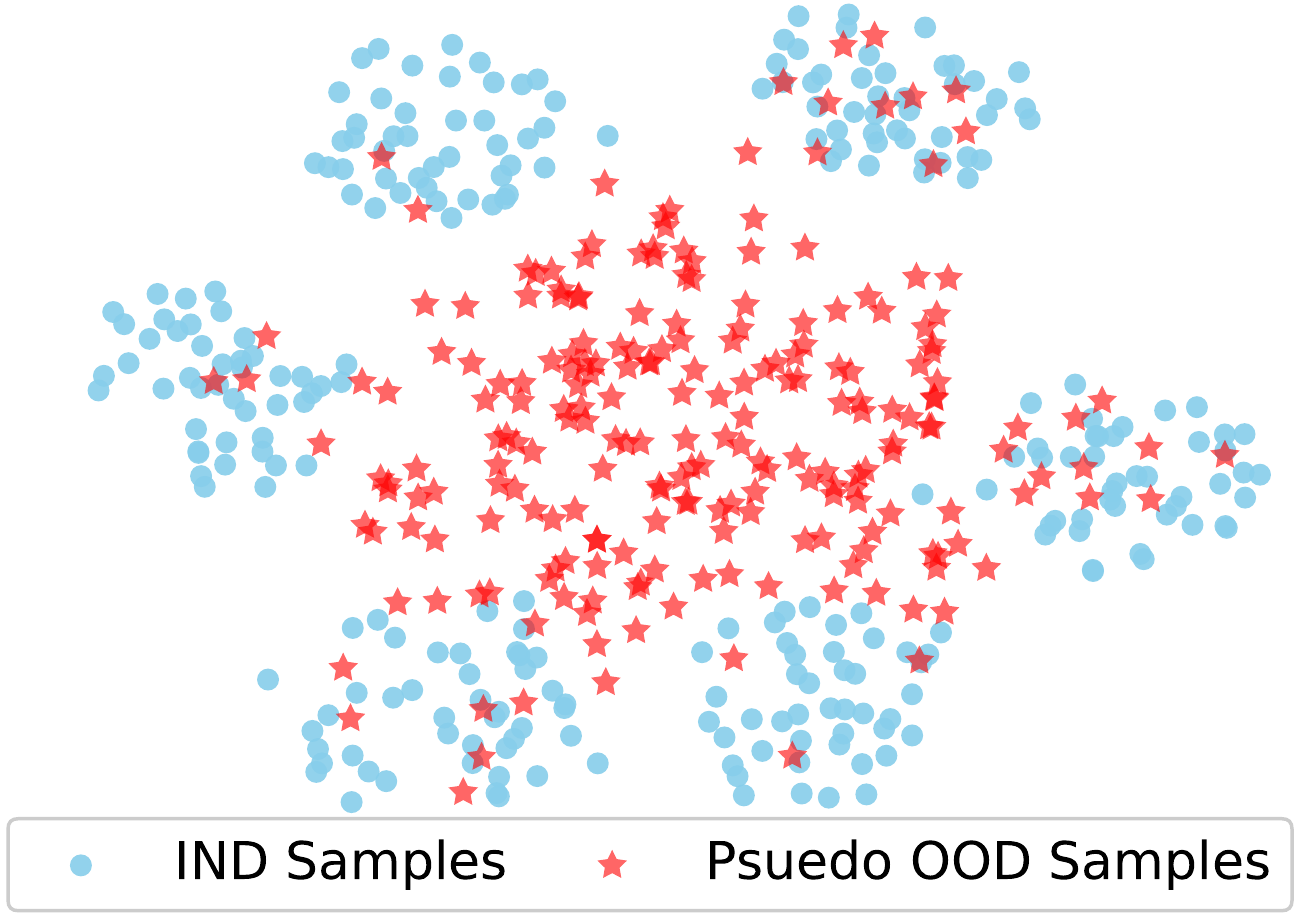} 
\caption{
  t-SNE visualization of pseudo OOD samples generated by feature mixup \citep{zhan2021out} on the CLINC150 dataset under the 25\% setting.
  Some pseudo OOD samples coincide with IND samples.}
\label{fig:CLINC150_0.25.pdf} 
\end{figure}

\begin{figure}[t] 
\centering 
\includegraphics[scale=0.41]{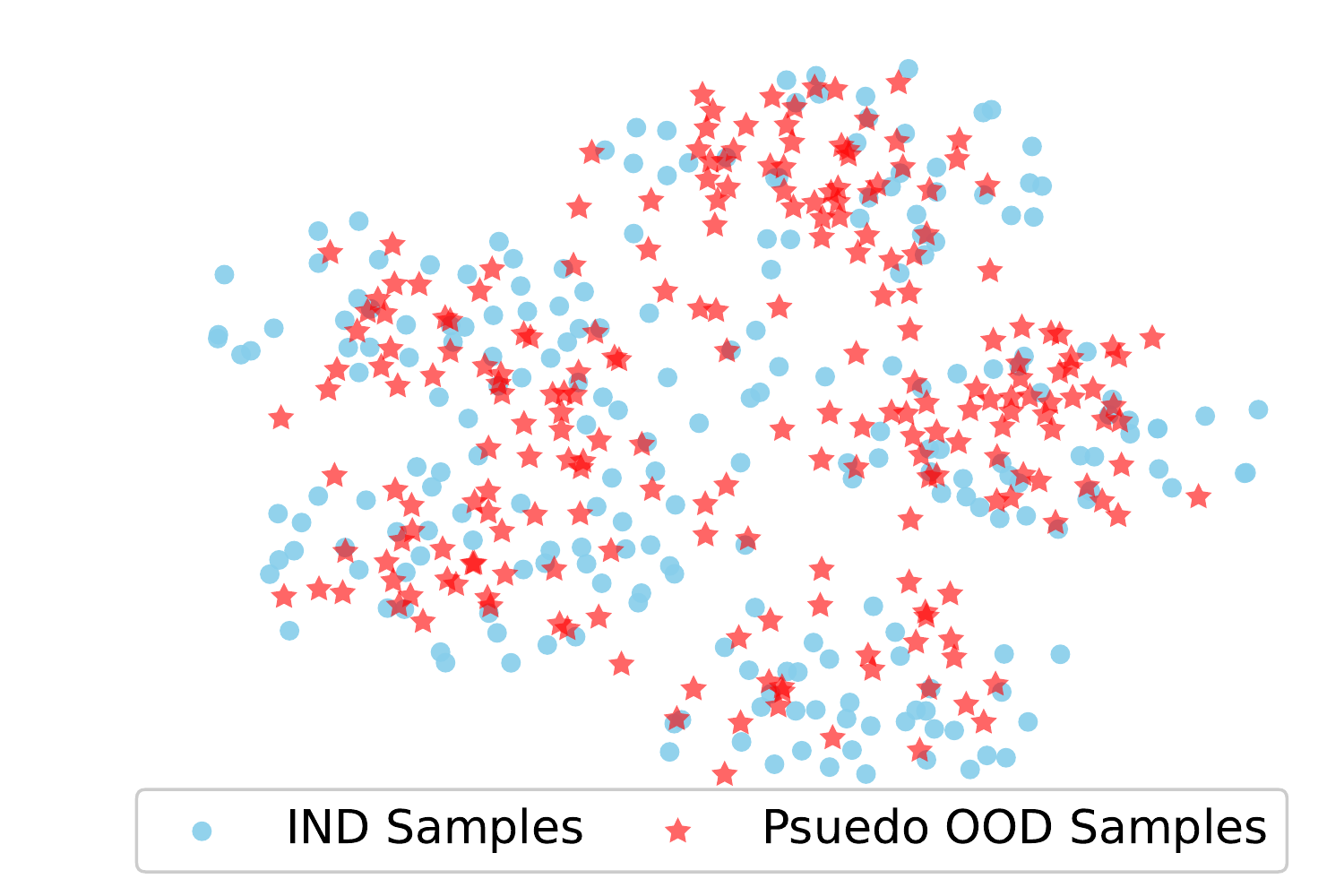} 
\caption{
  t-SNE visualization of pseudo OOD samples generated by feature mixup \citep{zhan2021out} on the StackOverflow dataset under the 25\% setting.
  Some pseudo OOD samples coincide with IND samples.}
\label{fig:stackoverflow_0.25.pdf} 
\end{figure}

Besides Figure \ref{fig:visulization_mixup}, we also visualize pseudo OOD samples produced by \citet{zhan2021out} on the CLINC150 \citep{larson2019evaluation} dataset (Figure \ref{fig:CLINC150_0.25.pdf}) and the StackOverflow \citep{xu2015short} dataset (Figure \ref{fig:stackoverflow_0.25.pdf}), when 25\% intents are randomly selected as IND intents. Specifically, \citet{zhan2021out} proposes to generate features of pseudo OOD samples by mixing up IND features. Pseudo OOD samples we demonstrate in this analysis are obtained using the code released by \citet{zhan2021out}. As shown in Figure \ref{fig:CLINC150_0.25.pdf} and \ref{fig:stackoverflow_0.25.pdf}, some pseudo OOD samples fall into the cluster of IND intents, and thus we argue it is improper to assign one-hot OOD labels to these samples.

Furthermore, we demonstrate more cases of pseudo OOD samples generated by the method proposed by \citet{shu2021odist} on the CLINC150 (Table \ref{tab:review_ood_click}), StackOverflow (Table \ref{tab:review_ood_stack}), and Banking (Table \ref{tab:review_ood_banking}) datasets. Specifically, these pseudo OOD samples are generated by replacing phrases in IND samples and use a pre-trained language model to filter these samples. As shown in Table \ref{tab:review_ood_click}, \ref{tab:review_ood_stack}, and \ref{tab:review_ood_banking}, some of the generated pseudo OOD samples carry IND intents since the replaced phrase may be an synonyms of the original phrase.

\begin{table*}
\begin{center}
\small
\begin{tabular}{  l|L{170pt} |L{170pt}} 
\toprule 
   \textbf{Original Intent} & \textbf{Original Utterance before Distortion} & \textbf{Generated Pseudo OOD Sample} \\ 
\midrule
  translate & how do you say hi in french & how do you spell hi in french \\ 
\midrule
  translate & how do you say please in arabic & how do you say please in Spanish? \\ 
\midrule
  transfer & transfer \$500 from my checking to my savings & transfer \$50 from my checking to my savings \\
\midrule
  transfer & can you transfer \$5 from savings to checking & can you transfer your \$5,000 from savings to checking \\
\midrule
  insurance\_change & what do i do for new insurance & what do you do for new insurance \\
\midrule
  insurance\_change & i would like to switch my insurance plan & i had to switch my insurance plan \\
\midrule
  travel\_alert & is it safe to travel to argentina & is it safe to travel to Europe? \\
\midrule
  travel\_alert & tell me about any travel alerts issued for germany & Read more about any travel alerts issued for germany \\
\midrule
  fun\_fact & can you tell me something i don't know about banks & can you tell me something i do n't know about you \\
\midrule
  fun\_fact & can you tell me fun facts about lighthouses & can you tell me fun facts about me? \\
\bottomrule
\end{tabular}
\end{center}
\caption{Case study of generated OOD samples with ODIST on the CLINC150 dataset.}
\label{tab:review_ood_click}
\end{table*}

\begin{table*}
\begin{center}
\small
\begin{tabular}{  l|L{170pt} |L{170pt}} 
\toprule
   \textbf{Original Intent} & \textbf{Original Utterance before Distortion} & \textbf{Generated Pseudo OOD Sample}  \\ 
\midrule
wordpress & Multiple loop working, function inside isn't &Multiple looping function inside is n't \\
\midrule
wordpress & Plugin to avoid share username in Wordpress & Plugin to share username in Wordpress \\
\midrule
visual-studio & Visual Studio Find and Replace Variables &  Visual Studio Find and Destroy \\
\midrule
visual-studio & Number of Classes in a Visual Studio Solution & Number of Users in a Visual Studio Solution \\
\midrule
svn & how to update a file in svn? & how do you open a file in svn? \\
\midrule
svn & Revert a whole directory in tortoise svn? & What is a whole directory in tortoise svn? \\
\midrule
spring & Problem with Autowiring \& No unique bean & Problem with Autowiring \& the unique bean \\
\midrule
spring & Creating temporyary JMS jms topic in Spring & Creating Your Own Home in Spring \\
\midrule
scala & Can Scala survive without corporate backing? & Can Scala get corporate backing? \\
\midrule
scala & What is are differences between Int and Integer in Scala? & What is are differences between Int and Int++ in Scala? \\
\bottomrule
\end{tabular}
\end{center}
\caption{Case study of generated OOD samples with ODIST on the StackOverflow dataset.}
\label{tab:review_ood_stack}
\end{table*}

\begin{table*}
\begin{center}
\small
\begin{tabular}{  L{80pt}|L{170pt} |L{170pt}} 
\toprule
   \textbf{Original Intent} & \textbf{Original Utterance before Distortion} & \textbf{Generated Pseudo OOD Sample}  \\ 
\midrule
   \shortstack{wrong\_exchange\_rate \\ \_for\_cash\_withdrawal} &  {While abroad I got cash, and a wrong exchange rate was applied.} &  {While abroad I got into an argument with a friend and a wrong exchange rate was applied.} \\ 
\midrule
   \shortstack{wrong\_exchange\_rate \\ \_for\_cash\_withdrawal} &  {This past holiday I made a withdraw at the ATM  machine, and it seems I've been charged too much.} &  {This morning I made a withdraw at the ATM machine,  and it seems I've been charged too much.}  \\
\midrule
   \shortstack{wrong\_amount\_of \\ \_cash\_received} &  {Where'd the rest of my cash go from the ATM} &  {Where 'd the rest of your cash go from the ATM}   \\
\midrule
   \shortstack{wrong\_amount\_of \\ \_cash\_received} &  {Why did I get less cash than what I asked in the  ATM?} &  {Why did I ask for less cash than what I asked in the  ATM?}   \\
\midrule
   {why\_verify\_identity} &  {I would like to know why so much identity  information is required?} &  {Who would like to know why so much identity  information is required?}   \\
\midrule
   {why\_verify\_identity} &  {Do i have to verify who I am?} & Why do i have to verify who I am?   \\
\midrule
   {verify\_top\_up} & How do I verify the top-up? & How do I use the top-up?    \\
\midrule
   {verify\_top\_up} & Please tell me how to verify my top up card. &  {Please check out this post for how to verify my top up  card.}    \\
\midrule
  verify\_my\_identity &  {Let me know what the steps for the identity checks  are} & We know what the steps for the identity checks are   \\
\midrule
  verify\_my\_identity & What documents do I need for the identity check? & What documents do we need for the identity check? \\
\bottomrule
\end{tabular}
\end{center}
\caption{Case study of generated OOD samples with ODIST on the Banking dataset.}
\label{tab:review_ood_banking}
\end{table*}

\section{More Implementation Details}\label{append:implementation_details}

We use the BERT model (\textit{bert-base-uncased}) provided in the Huggingface’s Transformers library \citep{wolf2020transformers} to implement $f$. Following \citep{zhang2021deep}, we add an averaging-pooling layer on top of BERT to obtain the representation of each input utterance. The projection head $h$, classification heads $g1$ and $g_2$ are implemented as a two-layer MLP with the LeakyReLU activation \cite{xu2020reluplex}, where the feature dimension is 1024 and the projection dimension is 128. Following \citep{zhan2021out}, We use AdamW \citep{kingma2014adam} to fine-tune BERT using a learning rate of 1e-5 and Adam \citep{wolf2019huggingface} to train the MLP heads using a learning rate of 1e-4 with early stopping. When learning the adaptive decision boundaries \citep{zhang2021deep}, the trained model is fixed and we used a learning rate of 0.05. We tried batch size of \{100, 200\} for IND samples and \{100, 500, 600, 800\} for OOD samples. All hyper-parameters are tuned according to the classification performance over the IND samples on the validation set. We find that $t=\tau=0.1$, $\alpha=0.11$ and $\beta=0.9$ work well with all datasets. We use a dropout rate 0.6 for the two classification heads. Each result is an average of 10 runs with different random seeds, and each run is stopped when we reach a plateau on the validation loss. ALL experiments were conducted in the Nvidia Tesla V100-SXM2 GPU with 32G graphical memory. Our model contains 112.06M model parameters.

\section{More Details about Baselines}\label{append:baseline}
Baseline results (MSP, DOC, OpenMax, LMCL, and ADB) are copied from \citep{zhang2021deep}. Results of the baseline Outlier are copied from \citep{zhan2021out}. Results of the baseline ODIST are copied from \citep{shu2021odist}. 
For above mentioned baselines, we also re-implement their methods using their release codes.
The results reproduced by our experiments match the results reported in their original paper.
So we copied the highest reported results of these baselines from previously published papers.
Significant tests between our method and all these baselines are carried out based on our implementations.
We get the baseline results (SCl and GOT) by running their released codes, and use 100 OOD samples in the validation to determine the thresholds for testing.

\section{Computational Cost Analysis}\label{append:comp}
We compare the training cost of our method when using one-hot labels or soft labels produced by ASoul.
Pseudo OOD samples used in this analysis are generated using the best performing method FM, and we use the CLINC150 dataset for this analysis. As shown in Table~\ref{tab:efficiency}, ASoul only introduces marginal parameter overhead for the projection head and the classification head. We can also observe that using ASoul only introduces little time overhead compared to the one-hot labeling approach.

\begin{table}[htbp]
\begin{center}
\small
\begin{tabular}{ l | l | l |l| l } 
\toprule
   Methods & \#para. & 25\%  & 50\%  & 75\% \\ 
\midrule
  FM+Onehot & 111.36M & 7.97s & 15.76s & 23.40s \\
  FM+Asoul & 112.06M & 8.57s & 16.85s & 25.06s \\
\bottomrule
\end{tabular}
\end{center}
\caption{Number of parameters (Million) and average training
time for each epoch (seconds) on the CLINC150 dataset.}
\label{tab:efficiency}
\end{table}

\section{Effect of Hyper-parameters}\label{append:hyper_sen}

We analyzed the most important hyper-parameters of ASoul: temperature $\tau$ in graph-based smoothing and dropout rate to the classification heads in co-training. We conduct experiments to show their effects on the CLINC150 dataset under the 25\% setting. We use the best performing pseudo OOD sample generation approach (i.e., FM) in this analysis.

\noindent\textbf{Temperature:} We set $\tau$ to \{0.1, 0.5, 1, 5, 10, 15, 20\} respectively, and demonstrate the performance change. Note that small $\tau$ makes distribution over the embedding graph more shape (concentrating on nearest neighbors), while large $\tau$ forms smooth distributions.

\begin{figure}[t] 
\centering 
\includegraphics[scale=0.35]{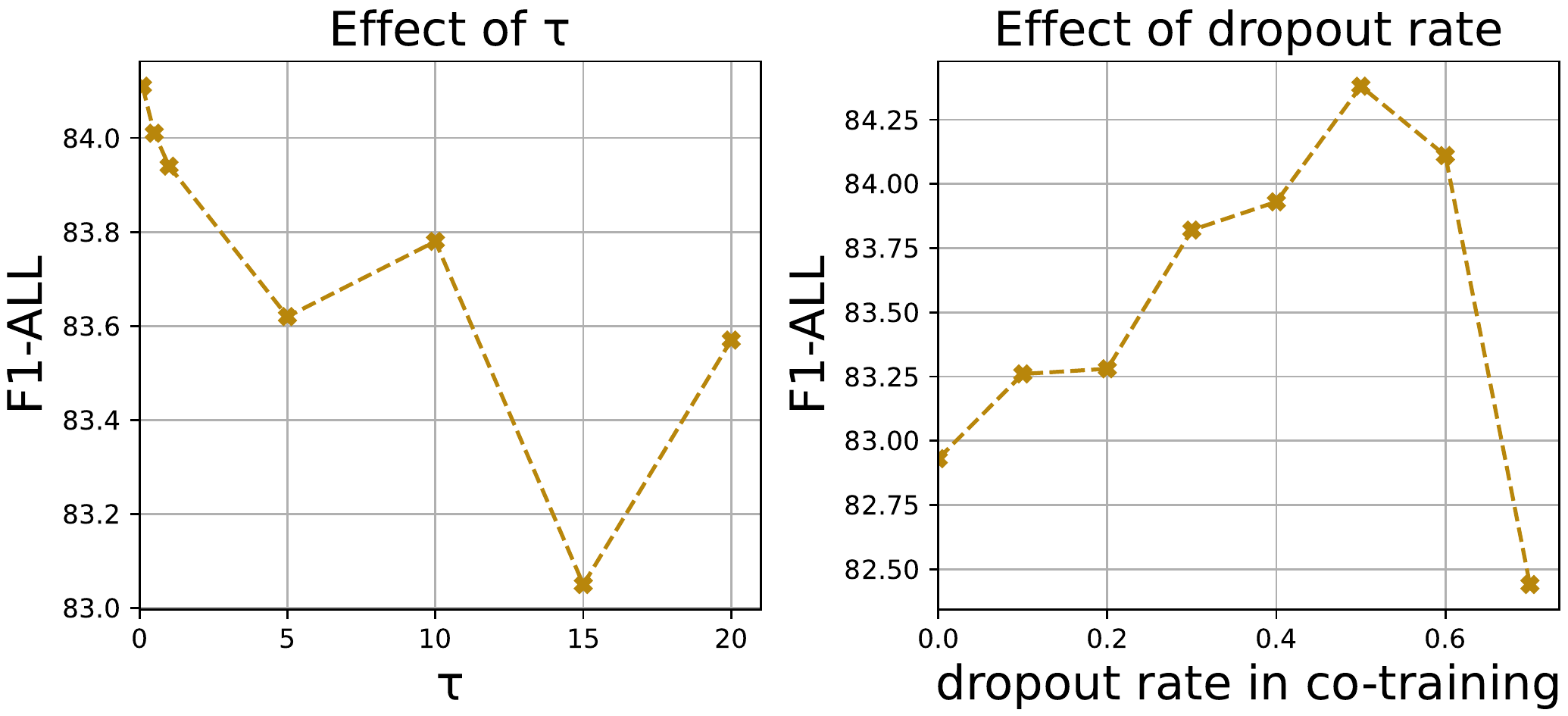} 
\caption{Effect of $\tau$ in graph-based smoothing (left) and dropout rate in co-training (right) with FM+ASoul on CLINC150 under the 25\% setting.}
\label{Fig.effect_of_paras} 
\end{figure}

Results are shown in Figure~\ref{Fig.effect_of_paras} (left). With the increase in temperature, the OOD detection performance tends to decrease. $\tau=0.1$ achieves the highest F1-ALL score of 84.11\%. This suggests that a small temperature makes ASoul focus more on neighbors and gain better performance.

\noindent\textbf{Dropout Rate:} We compare the performance of dropout rates to the classification heads by adjusting the rate from 0 to 0.7 with an interval of 0.1.

Results are shown in Figure~\ref{Fig.effect_of_paras} (right). The performance first increases and then decreases as the dropout rate increases. In the begging phase, using a high dropout rate introduces more diversity required by co-training, and thus the OOD detection performance improves. However, using a higher dropout rate introduces much noise to the co-training process, and thus downgrades the OOD detection performance.

\section{More Evaluation Metrics}\label{append:more_metrics}

We also calculate micro F1-scores over all intents (IND and OOD intents) for our best-performing method FM+ASoul and one of our strongest baselines Outlier on the CLINC150 dataset.
As shown in Table \ref{tab:more_metric}, FM+ASoul still outperforms the baseline on the micro F1-score.

\begin{table}[htbp]
\begin{center}
\small
\begin{tabular}{ l | l | l |l} 
\toprule
   Methods & 25\%  & 50\%  & 75\% \\ 
\midrule
  Outlier & 91.68 & 88.23 & 88.46 \\
  FM+Asoul & \textbf{93.30} & \textbf{90.36} & \textbf{89.98} \\
\bottomrule
\end{tabular}
\end{center}
\caption{ Performances of Outlier and FM+Asoul on the CLINC150 dataset under the metric of micro F1-score over all intents (IND and OOD intents) .}
\label{tab:more_metric}
\end{table}

\end{document}